\documentclass[runningheads]{llncs}

\usepackage{mathtools}      

\usepackage{graphicx}
\usepackage[dvipsnames]{xcolor}
\usepackage{cite}

\usepackage{amsfonts}       
\usepackage{booktabs}       
\usepackage{epsfig}         
\usepackage{float}          
\usepackage{graphicx}       
\usepackage{hyperref}       

\usepackage{microtype}      
\usepackage{nicefrac}       
\usepackage{subfig}
\usepackage{url}  

\usepackage{xspace}
\usepackage{amssymb}

\usepackage{multirow}
\usepackage{enumitem}

 \usepackage[ruled,vlined,linesnumbered,noresetcount]{algorithm2e}
 \usepackage{algpseudocode}

\usepackage{wrapfig}

\usepackage{thmtools,thm-restate}
\usepackage{cleveref} 

\usepackage[square,numbers]{natbib}

\usepackage{comment}

\allowdisplaybreaks

\long\def\ignorethis#1{}

\newcommand{\R}{\mathbb{R}}

\usepackage{bbm}
\usepackage{amsthm}

\newtheorem{Df}{Definition}
\newtheorem{Thm}{Theorem}

\begin{document}
\title{A Principled Approach to Data Valuation for Federated Learning}
%
%
\author{Tianhao Wang\inst{1}\and Johannes Rausch\inst{2}\and  Ce Zhang\inst{3}  \and Ruoxi Jia\inst{4}\and Dawn Song\inst{5}}
\authorrunning{Wang et al.}
\institute{Harvard University \email{tianhaowang@fas.harvard.edu}\and ETH Zurich \email{johannes.rausch@inf.ethz.ch} \and ETH Zurich \email{ce.zhang@inf.ethz.ch} \and Virginia Tech \email{ruoxijia@vt.edu} \and  UC Berkeley \email{dawnsong@gmail.com}}
%
\maketitle              
\begin{abstract}
Federated learning (FL) is a popular technique to train machine learning (ML) models on decentralized data sources. In order to sustain long-term participation of data owners, it is important to fairly appraise each data source and compensate data owners for their contribution to the training process. The Shapley value (SV) defines a unique payoff scheme that satisfies many desiderata for a data value notion. It has been increasingly used for valuing training data in centralized learning. However, computing the SV requires exhaustively evaluating the model performance on every subset of data sources, which incurs prohibitive communication cost in the federated setting. Besides, the canonical SV ignores the order of data sources during training, which conflicts with the sequential nature of FL. This paper proposes a variant of the SV amenable to FL, which we call the \emph{federated Shapley value}. The federated SV preserves the desirable properties of the canonical SV while it can be calculated without incurring extra communication cost and is also able to capture the effect of participation order on data value. We conduct a thorough empirical study of the federated SV on a range of tasks, including noisy label detection, adversarial participant detection, and data summarization on different benchmark datasets, and demonstrate that it can reflect the real utility of data sources for FL and has the potential to enhance system robustness, security, and efficiency. We
also report and analyze ``failure cases'' and hope to stimulate future research.

\end{abstract}

\section{Introduction}

Building high-quality ML models often involves gathering data from different sources. In practice, data often live in silos and agglomerating them may be intractable due to legal constraints or privacy concerns. FL is a promising paradigm which can obviate the need for centralized data. It directly learns from sequestered data sources by training local models on each data source and distilling them into a global federated model. FL has been used in applications such as keystroke prediction~\citep{hard2018federated}, hotword detection~\citep{leroy2019federated}, and medical research~\citep{medical}.

A fundamental question in FL is how to value each data source. FL makes use of data from different entities. In order to incentivize their participation, it is crucial to fairly appraise the data from different entities according to their contribution to the learning process. For example, FL has been applied to financial risk prediction for reinsurance~\citep{reinssurance}, where a number of insurance companies who may also be business competitors would train a model based on all of their data and and the resulting model will create certain profit. In order to prompt such collaboration, the companies need to concur with a scheme that can fairly divide the earnings generated by the federated model among them.

The SV has been proposed to value data in recent works~\citep{jia2019towards,jia2019efficient,ghorbani2019data}. The SV is a classic way in coopereative game theory to distribute total gains generated by the coalition of a set of players. One can formulate ML as a cooperative game between different data sources and then use the SV to value data. An important reason for employing the SV is that it \emph{uniquely} possesses a set of appealing properties desired by a data value notion: it ensures that (1) \emph{all} the gains of the model are distributed among data sources; (2) the values assigned to data owners accord with their actual contributions to the learning process; and (3) the value of data accumulates when used multiple times.

Despite the appealing properties of the SV, it cannot be directly applied to FL. By definition, the SV calculates the average contribution of a data source to every possible subset of other data sources. Thus, evaluating the SV incurs prohibitive communication cost when the data is decentralized. Moreover, the SV neglects the order of data sources, yet in FL the importance of a data source could depend on when it is used for training. For instance, in order to ensure convergence, the model updates are enforced to diminish over time (e.g., by using a decaying learning rate); therefore, intuitively, the data sources used toward the end of learning process could be less influential than those used earlier. Hence, a new, principled approach to valuing data for FL is needed.

In this paper, we propose \emph{the federated SV}, a variant of the SV designed to appraise decentralized, sequential data for FL. The federated SV can be determined from local model updates in each training iteration and therefore does not incur extra communication cost. It can also capture the effect of participation order on data value as it examines the performance improvement caused by each subset of players following the actual participation order in the learning process. Particularly, the federated SV preserves the desirable properties of the canonical SV. We present an efficient Monte Carlo method to compute the federated SV. Furthermore, we conduct a thorough empirical study on a range of tasks, including noisy label detection, adversarial participant detection, and data summarization on different benchmark datasets, and demonstrate that the federated SV can reflect the actual usefulness of data sources in FL.
We
also report and analyze cases in which the proposed federated SV can be further improved and hope to stimulate future research on this emerging topic.



\section{Related Work}
\label{sec:related_work}

Various data valuation schemes have been studied in the literature, and from a practitioner’s point of view they can be classified into query-based pricing that attaches prices to user-initiated queries~\citep{koutris2015query,upadhyaya2016price}; data attribute-based pricing that builds a price model depending on parameters such as data age and credibility using public price registries~\citep{heckman2015pricing}; and auction-based pricing that sets the price dynamically based on auctions~\citep{lee2010sell,mihailescu2010dynamic}. However, one common drawback of the existing strategies is that they cannot accommodate the unique properties of data as a commodity; for instance, the value of a data source depends on the downstream learning task and the other data sources used for solving the task.

The SV uniquely satisfies the properties desired by a data value notion. The use of the SV for pricing personal data can be traced back to~\citep{kleinberg2001value,chessa2017cooperative} in the context of marketing survey, collaborative filtering, recommendation systems, and networks.
Despite the desirable properties of the SV, computing the SV is known to be expensive. In its most general form, the SV can be $\mathsf{\#P}$-complete to compute~\citep{deng1994complexity}. The computational issue becomes even more serious when the SV is used to value training data for ML, because calculating it requires re-training models for many times. Most of the recent work on the SV-based data valuation has been focused on the centralized learning setting and improving its computational efficiency~\citep{jia2019towards,jia2019efficient,jia2019empirical,ghorbani2019data}.
Two important assumptions of the canonical SV are that the training performance on every combination of data points is measurable and that the performance does not depend on the order of training data. These two assumptions are plausible for centralized learning because the entire data is accessible to the coordinator and the data is often shuffled before being used for training. However, they are no longer valid for the federated setting. 

Existing work on pricing data in FL can be roughly categorized into two threads. One thread of work~\citep{kang2019incentive,kang2019incentivejoint} studies the mechanism design to incentivize participation given the disparity of data quality, communication bandwidth, and computational capability among different participants. In these works, the authors assume that the task publisher (i.e., the coordinator) has some prior knowledge about the data quality of a participant and design an optimal contract to maximize the utility of the coordinator subject to rationality constraints of individual participants. However, it remains a question how to precisely characterize data quality in FL. Another thread of work investigates the way to measure data quality and share the profit generated by the federated model according to the data quality measurement. \citet{wang2019measure} and \citet{song2019profit} apply the canonical SV to value each data source; however, as discussed earlier, the direct application of the SV is intractable in practice due to the decentralized data and conceptually flawed due to the sequential participation of participants. Recently, \citet{yu2020fairness} has studied at the intersection of two threads by proposing a fair profit sharing scheme while considering individual costs incurred by joining FL as well as the mismatch between contribution and payback time. Our work can be potentially integrated with their work to better characterize the data contribution.

\section{Data Valuation based on SV}
\label{sec:data_value}


Cooperative game theory studies the behaviors of coalitions formed by game players. Formally, a cooperative game is defined by a pair $(I,\nu)$, where $I=\{1,\ldots,N\}$ denotes the set of all players and $\nu:2^N\rightarrow \mathbb{R}$ is the utility function, which maps each possible coalition to a real number that describes the utility of a coalition, i.e., how much collective payoff a set of players can gain by forming the coalition.
One of the fundamental questions in cooperative game theory is to characterize the importance of each player to the overall cooperation. The SV~\cite{shapley1953value} is a classic method to distribute the total gains generated by the coalition of all players. The SV of player $i$ with respect to the utility function $\nu$ is defined as the average marginal contribution of $i$ to coalition $S$ over all $S\subseteq I\setminus \{i\}$: 
\begin{align}
\label{eqn:shapley_definition_no_order}
s^\nu_i = \frac{1}{N} \sum_{S\subseteq I\setminus\{i\}} \frac{1}{{ \binom{N-1}{|S|} }}
\big[\nu(S\cup \{i\})-\nu(S)\big]
\end{align}
We suppress the dependency on $\nu$ when the utility used is clear and use $s_i$ to represent the value allocated to player $i$.

The formula in (\ref{eqn:shapley_definition_no_order}) can also be stated in the equivalent form: 
\begin{align}
\label{eqn:shapley_definition_order}
    s_i = 
\frac{1}{N!}\sum_{\pi \in \Pi(I)}\big[ \nu(P_i^\pi\cup \{i\}) - \nu(P_i^\pi)\big]
\end{align}
where $\pi \in \Pi(I)$ is a permutation of players and $P_i^\pi$ is the set of players which precede player $i$ in $\pi$. Intuitively, imagine all players join a coalition in a random order, and that every player $i$ who has joined receives the marginal contribution that his participation would bring to those already in the coalition. To
calculate $s_i$, we average these contributions over all the possible orders.  

Applying these game theory concepts to data valuation, one can think of the players as data contributors and the utility function $\nu(S)$ as a performance measure of the model trained on the set of training data $S$. The SV of each data contributor thus measures its importance to learning an ML model. The following desirable properties that the SV \textit{uniquely} possesses motivate many prior works~\cite{kleinberg2001value,chessa2017cooperative,jia2018shapley,jia2019efficient,jia2019empirical,ghorbani2019data} to adopt it for data valuation.

\begin{enumerate}[leftmargin=*]
    \item {\bf Group Rationality}: The value of the model is completely distributed among all data contributors, i.e., $\nu(I) = \sum_{i\in I} s_i$.
    
    \item {\bf Fairness}: (1) Two data contributors who are identical with respect to what they contribute to a dataset's utility should have the same value. That is, if data contributor $i$ and $j$ are equivalent in the sense that $\nu(S\cup \{i\}) = \nu(S\cup \{j\}),\forall S\subseteq I\setminus \{i,j\}$, then $s_i=s_j$. (2) Data contributor with zero marginal contributions to all subsets of the dataset receive zero payoff, i.e.,
    if  $\nu(S\cup \{i\})=0$, $\forall S\subseteq I\setminus\{i\}$, then $s_i=0$. 

    \item {\bf Additivity}: The values under multiple utilities sum up to the value under a utility that is the sum of all these utilities: $s^{\nu_1}_i +s^{\nu_2}_i = s^{\nu_1+\nu_2}_i$ for $i\in I$.

\end{enumerate}

The \emph{group rationality} property states that any rational group of data contributors would expect to distribute the full yield of their coalition. The \emph{fairness} property requires that the names of the data contributors play no role in determining the value, which should be sensitive only to how the utility function responds to the presence of a seller. The \emph{additivity} property facilitates efficient value calculation when data are used for multiple applications, each of which is associated with a specific utility function. With additivity, one can compute value shares separately for each application and sum them up.

There are two assumptions underlying the definition of the SV:
\begin{enumerate}[leftmargin=*]
    \item \textbf{Combinatorially Evaluable Utility}: The utility function can be evaluated for every combination of players; 
    \item \textbf{Symmetric Utility}: The utility function does not depend on the order of the players.
\end{enumerate}
Both of the assumptions are plausible for centralized learning. Since the entire data is accessible to the coordinator, it is empowered to evaluate the model performance on the data from an arbitrary subset of contributors. Furthermore, in centralized learning, the data is often shuffled before being used for training. Hence, it is reasonable to consider the model performance to be independent the order of data points in the training set. In the next section, we will argue that these two assumption are no longer valid for FL and propose a variant of the SV amenable to the federated setting.

\section{Valuing Data for FL}
\label{sec:data_value_fed}

\subsection{Federated Shapley Value}

A typical FL process executes the following steps repeatedly until some stopping criterion is met: (1) The coordinator samples a subset of participants; (2) The selected participants download the current global model parameters from the coordinator; (3) Each selected participant locally computes an update to the model by training on the local data; (4) The coordinator collects an aggregate of the participant updates; (5) The coordinator locally updates the global model based on the aggregated update computed from the participants that participate in the current round. 

Let $I$ be the set of participants that participate in at least one round of the FL process. Our goal is to assign a real value to each participant in $I$ to measure its contribution to learning the model. Suppose the learning process lasts for $T$ rounds. Let the participants selected in round $t$ be $I_t$ and we have $I = I_1\cup \cdots \cup I_T$.


In FL, different participants contribute to the learning process at different time and the performance of the federated model depends on the participation order of participants. Clearly, the symmetric utility assumption of the SV does not hold. Moreover, FL is designed to maintain the confidentiality of participants' data and in each round, only a subset of participants are selected and upload their model updates. Hence, the coordinator can only know the model performance change caused by adding a participant's data into the subset of participants' data selected earlier. 
However, computing the SV requires the ability to evaluate the model performance change for every possible subset of participants. Unless the participants are able to bear considerable extra communication cost, the combinatorially evaluable utility assumption is invalid for FL. Hence, the SV cannot be used to value the data of different participants in the federated setting. 


We propose a variant of the SV amenable to the federated setting. The key idea is to characterize the aggregate value of the set of participants in the same round via the model performance change caused by the addition of their data and then use the SV to distribute the value of the set to each participant. We will call this variant \textit{the federated SV} and its formal definition is given below. We use $\nu(\cdot)$ to denote the utility function which maps any participants' data to a performance measure of the model trained on the data. Note that unlike in the canonical SV definition where $\nu(\cdot)$ takes a set as an input, the argument of $\nu(\cdot)$ is an ordered sequence. For instance, $U(A+B)$ means the utility of the model that is trained on A's data first, then B's data. Furthermore, let $I_{1:t-1}$ be a shorthand for $I_1+\cdots+I_{t-1}$ for $t\geq 2$ and $\emptyset$ for $t=1$.

\begin{Df}[The Federated Shapley  Value]
Let $I=\{1,\cdots,N\}$ denote the set of participants selected by the coordinator during a $T$-round FL process. Let $I_t$ be the set of participants selected in round $t$ and $I_t\subseteq I$. Then, the federated SV of participant $i$ at round $t$ is defined as
\begin{align}
\label{eqn:federated_shapley_per_round}
    s_t^\nu(i) = \frac{1}{|I_t|}\sum_{S\subseteq I_t\setminus \{i\}}\frac{1}{{  \binom{|I_t|-1}{|S|}  }}\big[\nu(I_{1:t-1} +  (S \cup \{i\})) - \nu(I_{1:t-1} +  S)\big] \; \text{if $i\in I_t$}
\end{align}
and $s_t(i) = 0$ otherwise. The federated SV takes the sum of the values of all rounds:
\begin{align}
\label{eqn:federated_shapley}
    s^\nu(i)  = \sum_{t=1}^T s_t(i)
\end{align}
\end{Df}
We will suppress the dependency of the federated SV $s^\nu(i)$ on $\nu$ whenever the underlying utility function is self-evident.

Due to the close relation between the canonical SV and the federated SV, one can expect that the federated variant will inherit the desirable properties of the canonical SV. Indeed, Theorem~\ref{thm:federated_shapley_property} shows that the federated SV preserves the group rationality, fairness, as well as additivity.

\begin{Thm}\label{thm:federated_shapley_property}
The federated SV defined in (\ref{eqn:federated_shapley}) uniquely possesses the following properties:
\begin{enumerate}
    \item Instantaneous group rationality: $\sum_{i\in I_t} s_t(i) = \nu(I_{1:t}) - \nu(I_{1:t-1})$.
    \item Fairness: (1) if $\nu(I_{1:t-1} + (S\cup\{i\})) = \nu(I_{1:t-1} + (S\cup \{j\}))$, $\forall S \subseteq I_t/\{i, j\}$ for some round $t$, then $s_t(i) = s_t(j)$. 
    (2) $\nu(I_{1:t-1} + (S\cup\{i\})) = \nu(I_{1:t-1} + S)$, $\forall S \subseteq I_t/\{i\}$ for some round $t$, then $s_t(i) = 0$. 
    \item Additivity: $s^{\nu_1+\nu_2}(i) = s^{\nu_1}(i) + s^{\nu_2}(i)$ for all $i\in I$.
\end{enumerate}
\end{Thm}

The proof of the theorem follows from the fact that the federated Shapley value calculates the Shapley value for the players selected in each round which distributes the performance difference from the previous round.

By aggregating the instantaneous group rationality equation over time, we see that the federated SV also satisfies the \emph{long-term} group rationality: 
\begin{align}
      \sum_{i=1}^N s(i) = U(I_1+\cdots+I_T)
\end{align}
The \emph{long-term} group rationality states that the set of players participates in a $T$-round FL process will divide up the final yield of their coalition.

\subsection{Estimating the Federated SV}

Similar to the canonical SV, computing the federated SV is expensive. Evaluating the exact federated SV involves computing the marginal utility of every participant to every subset of other participants selected in each round (see Eqn.~\ref{eqn:federated_shapley_per_round}). To evaluate $U(I_{1:t-1}+S)$, we need to update the global model trained on $I_{1:t-1}$ with the aggregate of the model updates from $S$ and calculate the updated model performance. The total complexity is $\mathcal{O}(T2^{m})$, where $m$ is the maximum number of participants selected per round. In this section, we present efficient algorithms to approximate the federated SV. 
We say that $\hat{s}\in \R^N$ is a $(\epsilon,\delta)$-approximation to the true SV $s=[s_1,\cdots,s_N]^T\in \R^N$ if $Pr[||\hat{s}_i-s_i||_\infty \leq \epsilon]\geq 1-\delta$. These algorithms utilize the existing approximation methods developed for the canonical SV~\cite{maleki2015addressing,jia2018shapley} to improve the efficiency of per-round federated SV calculation.



The idea of the first approximation algorithm is to treat the Shapley value of a participant as its expected contribution to the participants before it in a random permutation using Eqn.~\ref{eqn:shapley_definition_order} and use the sample average to approximate the expectation. We will call this algorithm \emph{permutation sampling-based approximation} hereinafter and the pseudocode is provided in Algorithm~\ref{alg:roundSV_sampling}. An application of Hoeffding bound indicates that to achieve $(\epsilon, \delta)$-approximation in each round of updating, the number of utility evaluations required for $T$ rounds is $Tm(\frac{2r^2}{\epsilon^2}) \log (\frac{2m}{\delta})$. 


The second approximation algorithm makes use of the group testing technique~\cite{jia2019towards} to estimate the per-round federated SV and we will call this algorithm \emph{group testing-based approximation}.  
In our scenario, each "test" corresponds to evaluating the utility of a subset of participant updates. 
The key idea of the algorithm is to intelligently design the sampling distribution of participants' updates so that we can calculate Shapely differences between the selected participants from the test results with high-probability bound on the error. Based on the result in~\cite{jia2019towards},
the number of tests required to estimate the Shapley differences up to $(\frac{\epsilon}{C_{\epsilon}}, 
\frac{\delta}{C_\delta})$ is 
$T_1 = \frac{4}{(1-q_{tot}^2)h(\frac{2\epsilon}{Zr C_{\epsilon} (1-q_{tot}^2)})} \log(\frac{C_{\delta}(m-1)}{2\delta})$, where $h(u)=(1+u)\log(1+u)-u$ and other variables are defined in Algorithm~\ref{alg:roundSV}.  
We can then take any participant $i$ and estimate the corresponding SV using the permutation sampling-based approximation, denote it as $s_*$. 
Then, the SV of all other $m-1$ users can be estimated using the estimated difference of the SV with participant $i$ (we choose the $m$th participant as the pivot participant in the pseudo-code in Algorithm \ref{alg:diffToSV}). 
The number of utility evaluation required for estimating $s_*$ up to $(\frac{(C_\epsilon-1)\epsilon}{C_\epsilon}, \frac{(C_\delta-1)\delta}{C_\delta})$ is $T_2 = \frac{4r^2 C_{\epsilon}^2}{(C_{\epsilon}-1)^2 \epsilon^2} \log (\frac{2C_{\delta}}{(C_{\delta}-1)\delta})$. $C_{\epsilon}, C_\delta$ are chosen so that $T_1 + T_2$ are minimized. 
Algorithm \ref{alg:federatedsv},
\ref{alg:roundSV_sampling},
\ref{alg:roundSV}, and \ref{alg:diffToSV} present the pseudo-code for both permutation sampling and group testing. 

If we treat $\epsilon, \delta$ as constant, $T_1+T_2 \sim \mathcal{O}((\log m)^2)$ while permutation sampling-based approximation is $\mathcal{O}(m\log m)$. Therefore, when the number of selected participants in each round is large, group testing-based approximation is significantly faster than permutation sampling-based one. One the other hand, when the number of selected participants is small, permutation sampling-based approximation is more preferable since its utility evaluation complexity tends to have a smaller constant.

\newcommand{\algwidth}{10}

\begin{algorithm}
\SetAlgoLined
\SetKwInOut{Input}{input}
\SetKwInOut{Output}{output}
\Input{$N$ - available participants, 
$C$ - fraction of selected participants in each round,
\texttt{ParticipantUpdate} - function for participant's local update, e.g. SGD}
\Output{$(\hat s_1, \hat s_2, \dots, \hat s_N)$ - estimated SV for all participants}

Initialize global model $w_0$; initialize $\hat S_i \leftarrow 0$ for $i=1 \dots N$. 

\For{each round $t=0, 1, \dots,$}{
    $m \leftarrow \max(C\cdot N)$
    
    $C_t \leftarrow$ random set of $m$ participants
    
    \For{each participant $k \in C_t$ in parallel}{
    $w_{t+1}^k \leftarrow \texttt{ParticipantUpdate}(k, w_t)$
    }
    
    $\hat S[C_t] \leftarrow \hat S[C_t] + \texttt{RoundSVEstimation}(\{w_{t+1}^k\}, w_t)$
    
    $w_{t+1} \leftarrow \frac{1}{m} \sum_{k=1}^m w_{t+1}^k$
}
\Return{$\hat S$}
 \caption{Federated SV Estimation.}
 \label{alg:federatedsv}
\end{algorithm}

\begin{algorithm}
\SetAlgoLined
\SetKwInOut{Input}{input}
\SetKwInOut{Output}{output}
\Input{$\{w_{t+1}^{i}\}_{i=1}^m$ - selected participants' updates in round $t+1$, 
$w_t$ - global model in round $t$,
$U(\cdot) \in [0, r]$ - utility function, $\epsilon, \delta$ - approximation parameter}
\Output{ $(\hat s_1, \dots, \hat s_m)$ - estimated SV for selected participants}

$T \leftarrow \frac{2r^2}{\epsilon^2} \log(\frac{2m}{\delta})$

$U_{prev} \leftarrow U(w_t)$

Initialize $\hat s_i \leftarrow 0$ for $i = 1 \dots m$. 

\For{$t=1\dots T$}{
    
    Uniformly sample a permutation $S$ of set $\{w_{t+1}^i\}_{i=1}^m$.
    
    \For{$i=1\dots m$}{
    $\hat s_i \leftarrow 
    \hat s_i + (U(w_t + S[:i])-U_{prev})$
    
    $U_{prev} \leftarrow 
    U(w_t + S[:i])$
    }
}

$\hat s \leftarrow \hat s / T$

\Return{$\hat s$}

 \caption{\texttt{RoundSVEstimation} estimates the SV $s_t$ for selected participants at round $t$ using permutation sampling}
 \label{alg:roundSV_sampling}
\end{algorithm}

\begin{algorithm}
\SetAlgoLined
\SetKwInOut{Input}{input}
\SetKwInOut{Output}{output}
\Input{$\{w_{t+1}^{i}\}_{i=1}^m$ - selected participants' updates in round $t+1$, 
$w_t$ - global model in round $t$,
$U(\cdot) \in [0, r]$ - utility function, $\epsilon, \delta$ - approximation parameter, $C_{\epsilon}, C_{\delta}$ - tradeoff parameters}
\Output{ $(\hat s_1, \dots, \hat s_m)$ - estimated SV for selected participants}

$U_{tot} \leftarrow U(w_t + \{w_{t+1}^{i}\}_{i=1}^m)$

$Z \leftarrow 2\sum_{k=1}^{m-1} \frac{1}{k}$

$q(k) \leftarrow \frac{1}{k}(\frac{1}{k}+\frac{1}{m-k})$ for $k=1\dots m-1$. 

$q_{tot} \leftarrow \frac{m-2}{m}q(1) + \sum_{k=2}^{m-1}q(k)[1+\frac{2k(k-m)}{m(m-1)}]$

$T \leftarrow \frac{4}{(1-q_{tot}^2)h(\frac{2\epsilon}{Zr C_{\epsilon} (1-q_{tot}^2)})} \log(\frac{C_{\delta}(m-1)}{2\delta})$

Initialize $(a)_{ti} \leftarrow 0$, $t=1 \dots T$, $i = 1 \dots m$. 

\For{$t=1\dots T$}{
    Draw $k_t \sim q(k)$
    
    Uniformly sample a length-$k_t$ sequence $S$ from $\{1, \dots, m\}$.
    
    $a_{ti} \leftarrow 1$ for all $i \in S$
    
    $B_t \leftarrow U(w_t + S)$
}

$C_{ij} \leftarrow \frac{Z}{T} \sum_{t=1}^T B_t (A_{ti}-A_{tj})$ for $i=1 \dots m, j=1 \dots m$. 

$\hat s \leftarrow \texttt{DiffToSV}(C_{ij})$

 \caption{\texttt{RoundSVEstimation} estimates the SV $s_t$ for selected participants at round $t$ using group testing}
 \label{alg:roundSV}
\end{algorithm}

\begin{algorithm}
\SetAlgoLined

$T \leftarrow \frac{4r^2 C_{\epsilon}^2}{(C_{\epsilon}-1)^2 \epsilon^2} \log (\frac{2C_{\delta}}{(C_{\delta}-1)\delta})$

$\hat s_* \leftarrow 0$

\For{$t=1 \dots T$}{
    Uniformly sample a subset $S$ from $\{1, \dots, m-1\}$
    
    $\hat s_* \leftarrow \hat s_* +
    \frac{1}{T}(U(w_t + S\cup \{m\})-U(w_t + S))$
}

\For{i=1\dots m}{
    $\hat s_i = \hat s_* + C_{im}$
}

\Return{$\hat s = (\hat s_1, \dots, \hat s_m)$}

 \caption{\texttt{DiffToSV} recovers the SV $s_t$ for selected participants at round $t$ from the estimated difference $C_{ij}$. }
 \label{alg:diffToSV}
\end{algorithm}

\section{Empirical Study}
\label{sec:exp}

In this section, we conduct the first empirical evaluation on a range of real-world FL tasks with different datasets to study whether  the proposed data value notion can reflect the real utility of data.
The tasks include noisy data detection, adversarial participant removal and data summarization. We would expect that a good data value notion will assign low value to participants with noisy, adversarial, and low-quality data, which will in turn help us remove those participants.

\subsection{Baseline Approaches}

We will compare the federated SV with the following two baselines.

\textbf{Federated Leave-One-Out}
One natural way to assign the contribution to a participant update $i$ at round $t$ is by calculating the model performance change when the participant is removed from the set of participants selected at round $t$, i.e., $loo_{t}(i) = U(I_{1:t}) - U(I_{1:t-1}+I_t / \{i\})$,
and $loo_t(i)=0$ if participant $i$ is not selected in round $t$. The Leave-One-Out (LOO) value for FL takes the sum of the LOO values of all rounds: $loo(i) = \sum_{t=1}^T loo_t(i)$.

\textbf{Random}
The random baseline does not differentiate between different participants' contribution and just randomly selects participants to perform a given task.

In the figures, we will use Fed. LOO and Fed. SV to denote federated leave-one-out and federated Shapley Value approach, respectively.

\subsection{Experiment Setting}

For each task, we perform experiments on the MNIST~\citep{mnist} as well as the CIFAR10 dataset~\citep{cifar}.
Following~\citep{mcmahan2017communication}, we study two ways of partitioning the MNIST data over
participants: \textbf{IID}, where the data is shuffled, and then partitioned
into 100 participants each receiving 600 examples, and \textbf{Non-IID},
where we first sort the data by digit label, divide it into 200
shards of size 300, and assign each of 100 participants 2 shards.
For MNIST, we train a simple multilayer-perceptron (MLP) with 2-hidden layers with 200 neurons in each layer and ReLu activations as well as a simple CNN.
For all experiments on CIFAR10, we train a CNN with two 5x5 convolution layers (the first with 32 channels, the second with 64, each followed by 2x2 max pooling), a fully connected layer with 512 neurons with ReLu activation, and a final softmax output layer.
In each round of training, we randomly select 10 participants out of 100, unless otherwise specified. 
We run 25 rounds for training on MNIST, achieving up to 97\% and 92\% global model accuracy for the IID and the non-IID setting, respectively.
For CIFAR10, we run up to 50 to 200 rounds of training. 
We achieve up 77\% and 70\% test accuracy in IID and non-IID setting, respectively, for 200 rounds of training. 
As a side note, the state-of-the-art models in~\cite{kolesnikov2019big} can achieve test accuracy of 99.4\% for CIFAR10; nevertheless, our goal is to evaluate the proposed data value notion rather than achieving the best possible accuracy. We use the permutation sampling approach in Algorithm \ref{alg:roundSV_sampling} to estimate the Shapley value in all experiments since the number of participants is small.

\subsection{Noisy Label Detection}

Labels in the real world are often noisy due to  annotators’ varying skill-levels, biases or malicious tampering. 
We show that the proposed data value notion can help removing the noisy participants. The key idea is to rank the participants according to their data value, and drop the participants with the lowest values.

\begin{figure}[ht]
  \centering
  \subfloat[MNIST]{
    \includegraphics[width= 0.4 \linewidth]{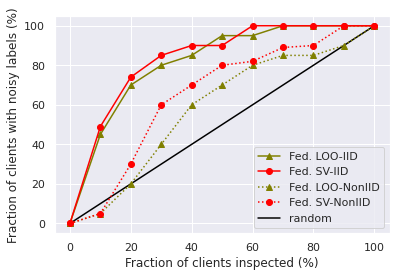}
    \label{fig:noisy-mnist}
  }
  \subfloat[CIFAR10]{
    \includegraphics[width=0.4 \linewidth]{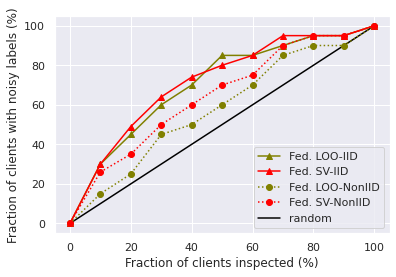}
    \label{fig:noisy-cifar}
  } \\
  \subfloat[MNIST - norm.]{
    \includegraphics[width=0.4 \linewidth]{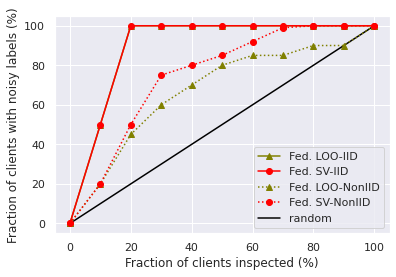}
    \label{fig:noisy-mnist-weighted}
  }
  \subfloat[CIFAR10 - norm.]{
    \includegraphics[width=0.4 \linewidth]{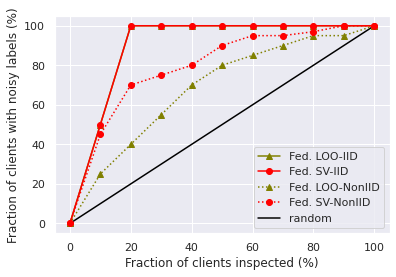}
    \label{fig:noisy-cifar-weighted}
  }
\caption{Experiment results of (a) (b) noisy label detection; (c) (d) noisy label detection with normalized federated LOO/SV. }
\label{fig:noisy}
\end{figure}

%
We set 20 participants' local data to be noisy where noise flipping ratio is 10\% for MNIST, and 3\% for CIFAR10. 
The performances of different data value measures are illustrated in Figure \ref{fig:noisy-mnist} and \ref{fig:noisy-cifar}. 
We inspect the label of participant's local training instances that have the lowest scores, and plot the change of the fraction of detected noisy participants with the fraction of the inspected participants.
We can see that when the training data is partitioned in IID setting, federated LOO and federated SV perform similarly. However, in the Non-IID setting, the federated SV outperforms federated LOO. We conjecture that this is because for Non-IID participants, the trained local models tend to overfit, diverge from the global model, and exhibit low accuracy. In comparison with the federated SV, federated LOO only computes the marginal contribution of a participant to the largest subset of other selected participants and therefore the noisy participants are harder to be identified by federated LOO.

\begin{figure}[ht]
  \centering
  \subfloat[Norm]{
    \includegraphics[width=.3\linewidth,height=.2\linewidth]{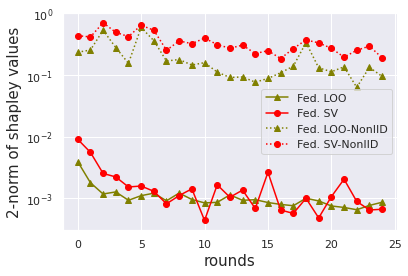}
    \label{fig:contribution}
  }
  \subfloat[MNIST]{
    \includegraphics[width=.25\linewidth,height=.2\linewidth]{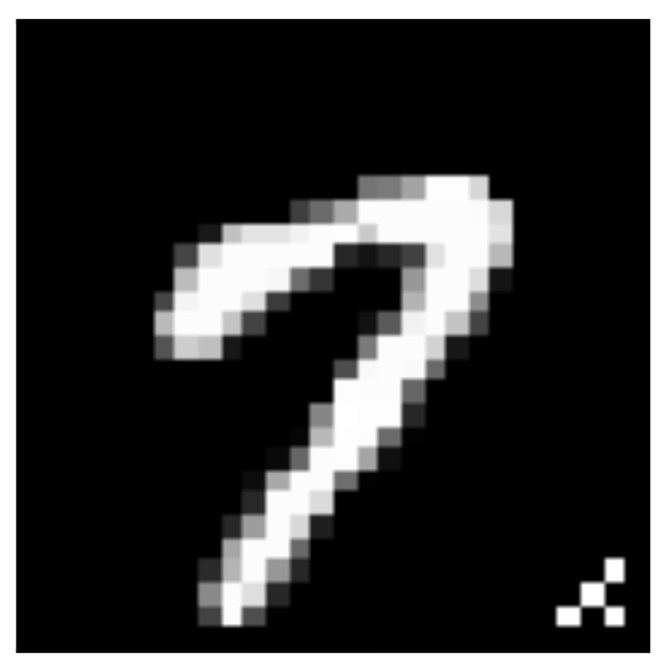}
    \label{fig:mnist-backdoor-ex}
  }
  \subfloat[CIFAR10]{
    \includegraphics[width=.25\linewidth,
    height=.2\linewidth]{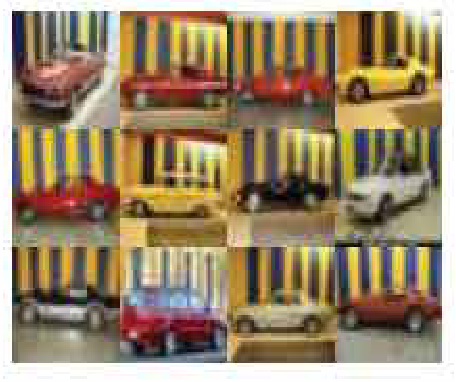}
    \label{fig:cifar-backdoor-ex}
  }
\caption{(a) Norm of Contribution varies with different rounds for MNIST-IID; (b) (c)  Illustrations of Backdoor Image}
\label{fig:backdoor-ex}
\end{figure}

We also find that, with the number of training rounds increases, the total contribution of participants in each round will decrease, as shown in Figure \ref{fig:contribution}. This makes sense since the federated SV satisfies instantaneous group rationality in Theorem \ref{thm:federated_shapley_property}, and the improvement of global model's utility will slowdown when it is close to convergence. That is, it is relatively easy to improve the global model's utility in earlier rounds, while harder to further improve the utility in later rounds. Hence, the contribution of participants selected in early rounds is inflated.
This inspires us to consider a variant of data value measures, which normalize the per-round data values by their norms and then aggregate them across all rounds.
The performance of noisy label detection with the normalized versions of federated SV and federated LOO is shown in Figure \ref{fig:noisy-mnist-weighted} and \ref{fig:noisy-cifar-weighted}. 
As we can see, it is much easier to separate noisy participants from benign participants with the normalized version of data value notions. 
However, the normalized federated SV no longer preserves the group rationality and additivity property. 
We leave developing more detailed analysis of different variants of data value as future work.

\newcommand{\imagewidth}{0.45}

\subsection{Backdoor Attack Detection}

Motivated by privacy concerns, in FL, the coordinator is designed to have no visibility into a participant’s local data and training process.
This lack of transparency in the agent updates can be exploited so that an adversary controlling a small number of malicious participants can perform a \emph{backdoor attack}. 
The adversary's objective is to cause the jointly trained global model to misclassify a set of chosen inputs, i.e. it seeks to poison the global model in a targeted manner, while also ensures that the global model has a good performance on the clean test data. We focus on backdoor attacks based on the model replacement paradigm proposed by \cite{bagdasaryan2020backdoor}. 


For CIFAR10, following the settings in \cite{bagdasaryan2020backdoor}, we choose the feature of vertically stripped walls in the background (12 images) as the backdoor. 
For MNIST, 
we implement pixel-pattern backdoor attack in \cite{gu2019badnets}. 
We set the ratio of the participants controlled by the adversary to be 30\%.
We mix backdoor images with benign images in every training batch (20 backdoor images per batch of size 64) for compromised participants, following the settings in \cite{bagdasaryan2020backdoor}. 

In Figure \ref{fig:mnist-backdoor-plot} and \ref{fig:cifar-backdoor-plot}, we show the success rate of backdoor detection with respect to the fraction of checked participants. Both of the figures indicate that federated SV is a more effective than federated LOO for detecting compromised participants. 
In the Non-IID setting, both compromised participants and some benign participants tend to have low contribution on the main task performance, which makes the compromised participants more difficult to be identified by the low data values. Hence, we also test the performance of normalized version of federated SV/LOO for this task and Figure~\ref{fig:mnist-backdoor-weighted-plot} and~\ref{fig:cifar-backdoor-weighted-plot} show that the performance improves a lot compared with the original definitions.

\subsection{Data Summarization} 

In our data summarization experiments, we investigate whether the federated SV can facilitate federated training by identifying the most valuable participants. Per communication round, a percentage of the selected participants is ignored for the update of the global model. We use data value measures to dismiss participants that are expected to contribute the least to the model accuracy. The data values are calculated on a separate validation set, which contains 1000 and 800 random samples for MNIST and CIFAR10, respectively.
During each communication round of FL, we compute the data value summands. After training has finished, we compute the total data value. 

We then repeat training, while maintaining an identical selection of participants per round. During each round, we dismiss a certain fraction $q \in [0, 0.1, \ldots, 0.9]$ of the selected participants. We compute and average the results for the random baseline three times per run.

We train a small CNN model on the MNIST dataset.  
The CNN consists of two 5x5 convolution layers, each followed with 2x2 max pooling and ReLu activations. Two fully connected layers (input, hidden and output dimensions of 320, 50, 10, respectively) with intermediate ReLu activation follow the second convolution layer. We apply dropout on the second convolution and first fully connected layer. 
For CIFAR10, we operate on 1000-dimensional feature vectors extracted with an imagenet-pretrained MobileNet v2 mode.\footnote{We use preprocessing and the pretrained model as provided by PyTorch Hub} We train a MLP with 2-hidden layers with 1000 neurons in each layer and ReLu activations.

We evaluate our algorithm for FL of 10 rounds on MNIST and 100 rounds on CIFAR10. The results of our experiments are shown in Figure \ref{fig:mnist_summarization_10rounds_and_25rounds}.
For the MNIST IID case, the federated SV approach outperforms both baselines. While it also consistently outperforms the random baseline in the non-IID setting, federated LOO achieves higher test accuracies for lower fractions of dismissed samples. Here, analysis of the federated SV per participant shows that it tends to be higher for participants that are selected throughout the FL. Furthermore, we observe that participants that were sampled few times also are more likely to have a negative federated SV, compared to the IID setting. We hypothesize that this bias negatively affects the performance of the federated SV-based summarization in the non-IID setting. 

We also observe that both federated SV and LOO perform worse on the CIFAR10 dataset summarization. We hypothesize that selection of good participant subsets is more effective on the MNIST dataset, as it contains a larger portion of redundant samples. Consequently, valuable information is less likely to be lost by dismissal of a fraction of participants.

\section{Conclusion}
\label{sec:conclusion}

This paper proposes the federated SV, a principled notion to value data for the process of FL. 
The federated SV uniquely possesses the properties desired by a data value notion, including group rationality, fairness, and additivity, while enjoying communication-efficient calculation and being able to capture the effect of participant participation order on the data value. We present algorithms to approximate the federated SV and these algorithms are significantly more efficient than the exact algorithm when the number of participants is large. Finally, we demonstrate that the federated SV can reflect the actual utility of data sources through a range of tasks, including noisy label detection, adversarial participant detection, and data summarization. 


\newcommand{\height}{3.2cm}

\begin{figure}[ht]
  \centering
  \subfloat[MNIST]{
    \includegraphics[width= 0.4 \linewidth, height=\height]{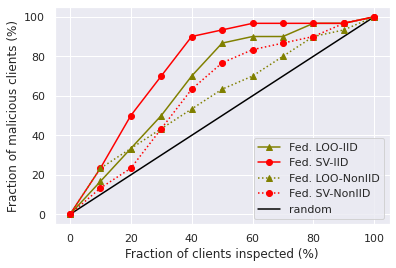}
    \label{fig:mnist-backdoor-plot}
  }
  \subfloat[CIFAR10]{
    \includegraphics[width=0.4 \linewidth, height=\height]{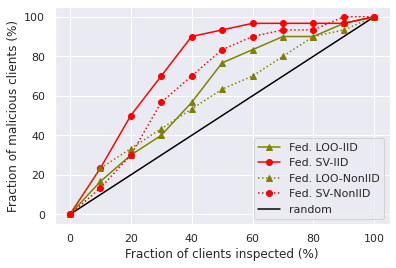}
    \label{fig:cifar-backdoor-plot}
  } \\
  \subfloat[MNIST - norm.]{
    \includegraphics[width=0.4 \linewidth, height=\height]{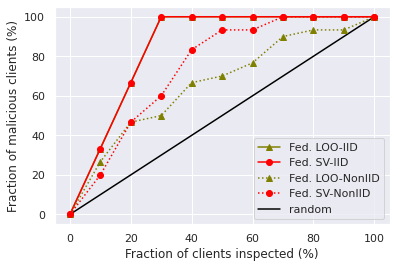}
    \label{fig:mnist-backdoor-weighted-plot}
  }
  \subfloat[CIFAR10 - norm.]{
    \includegraphics[width=0.4 \linewidth, height=\height]{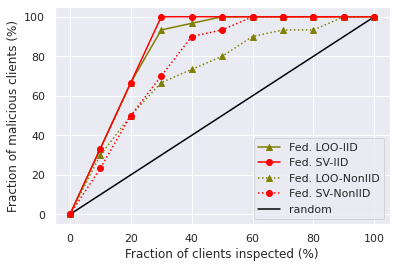}
    \label{fig:cifar-backdoor-weighted-plot}
  }
\caption{Experiment results of (a) (b) backdoor detection; (c) (d) backdoor detection with normalized LOO/SV. }
\label{fig:backdoor-plot}
\end{figure}

\begin{figure}[ht]
  \centering
  \subfloat[MNIST, IID, $T=10$]{
    \includegraphics[trim=0cm 0cm 0cm 1.2cm, clip=true,width=0.4 \linewidth, height=\height]{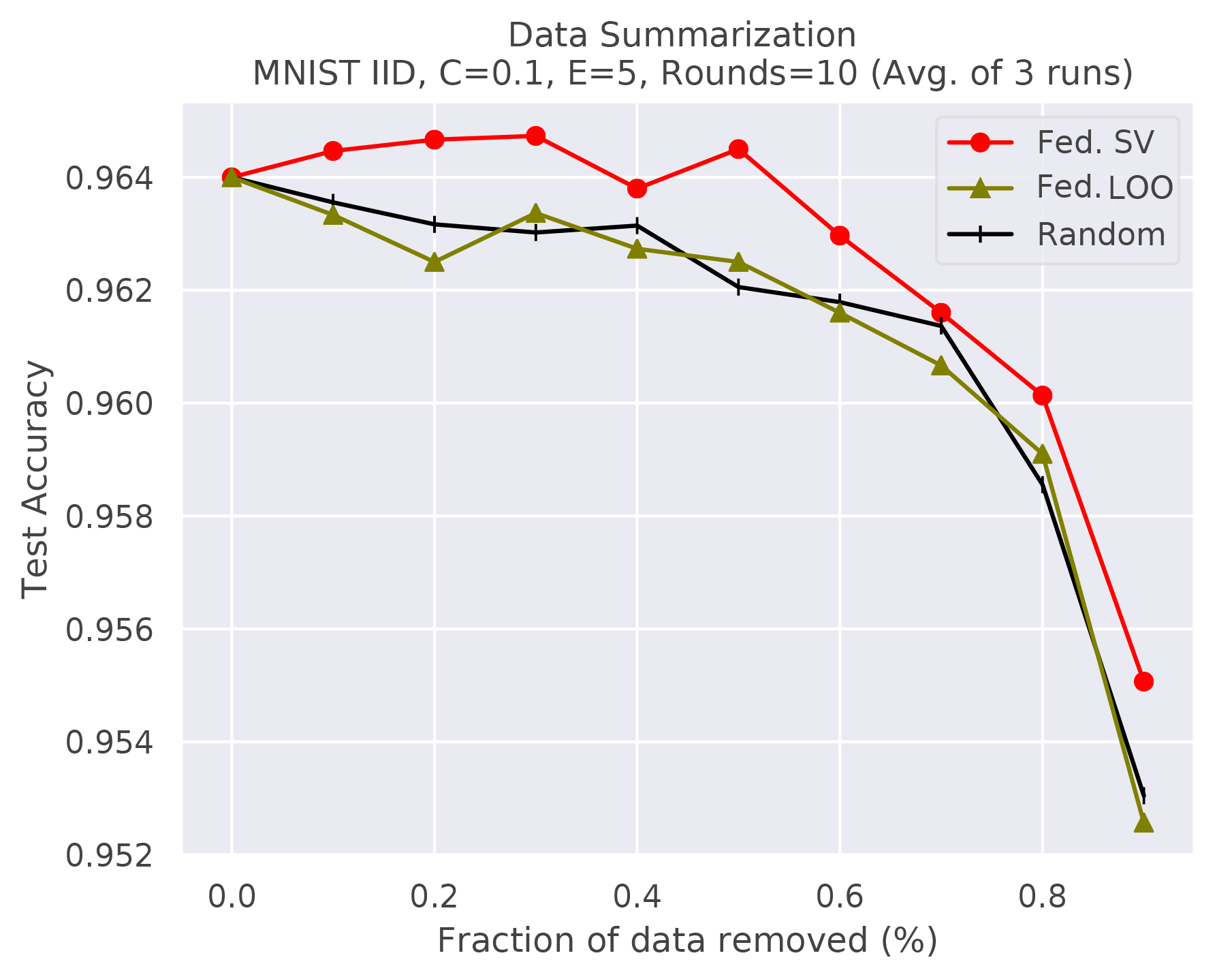}
  }
  \subfloat[MNIST, non-IID, $T=10$]{
    \includegraphics[trim=0cm 0cm 0cm 1.2cm, clip=true,width=0.4 \linewidth, height=\height]{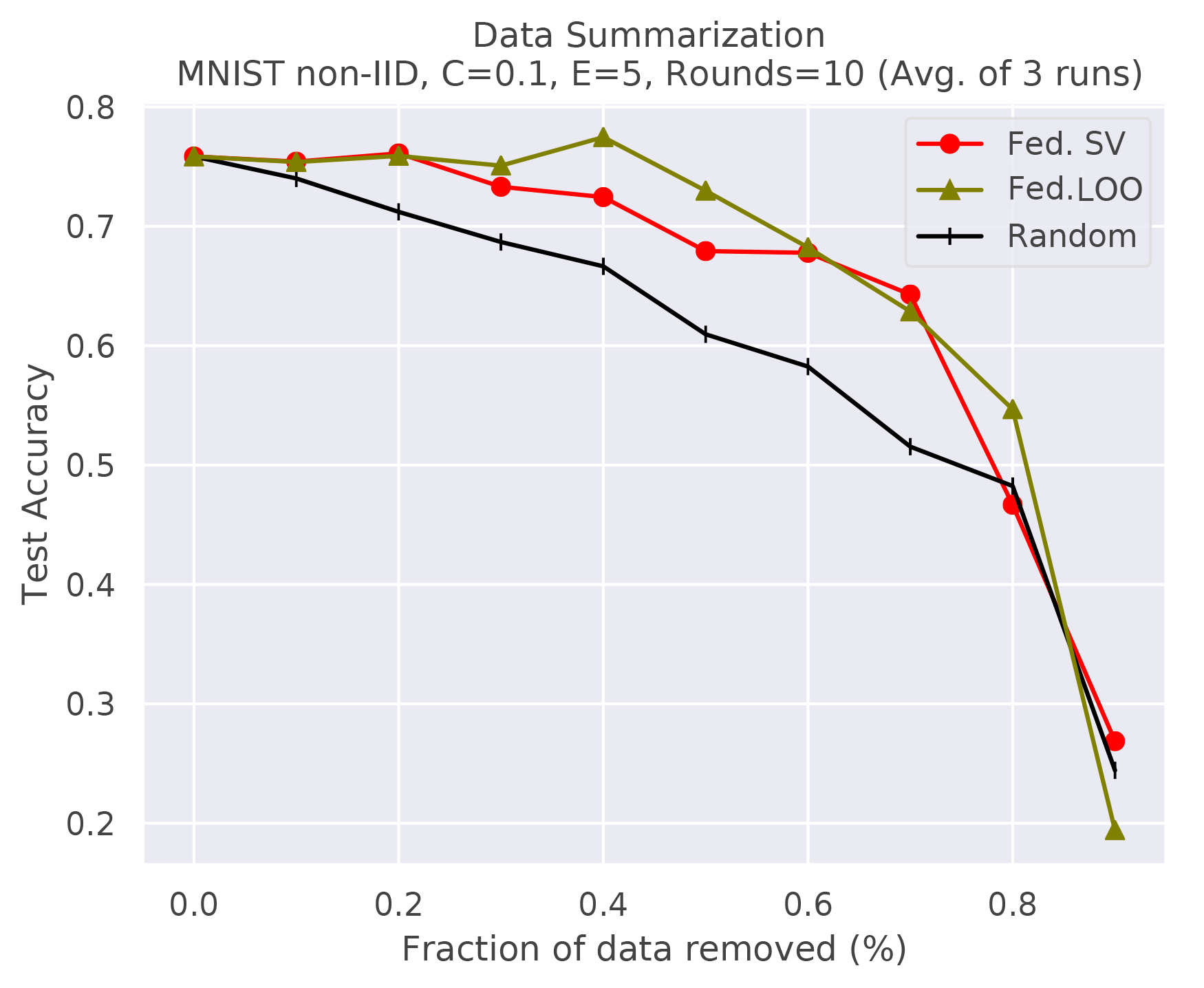}
  }\\
  \subfloat[CIFAR, IID, $T=100$]{
    \includegraphics[trim=0cm 0cm 0cm 1.2cm, clip=true,width=0.4 \linewidth, height=\height]{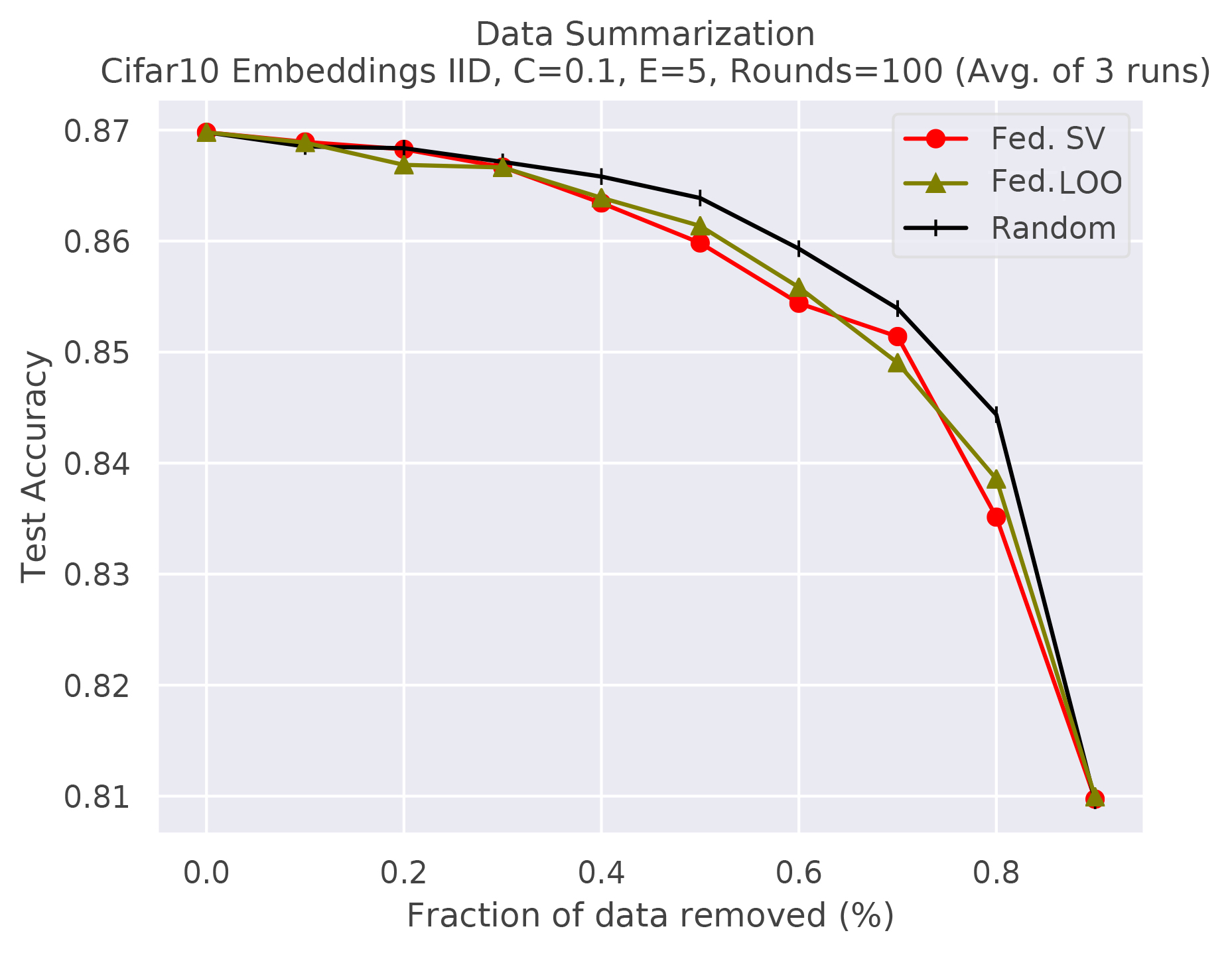}
  }
  \subfloat[CIFAR, non-IID, $T=100$]{
    \includegraphics[trim=0cm 0cm 0cm 1.2cm, clip=true,width=0.4 \linewidth, height=\height]{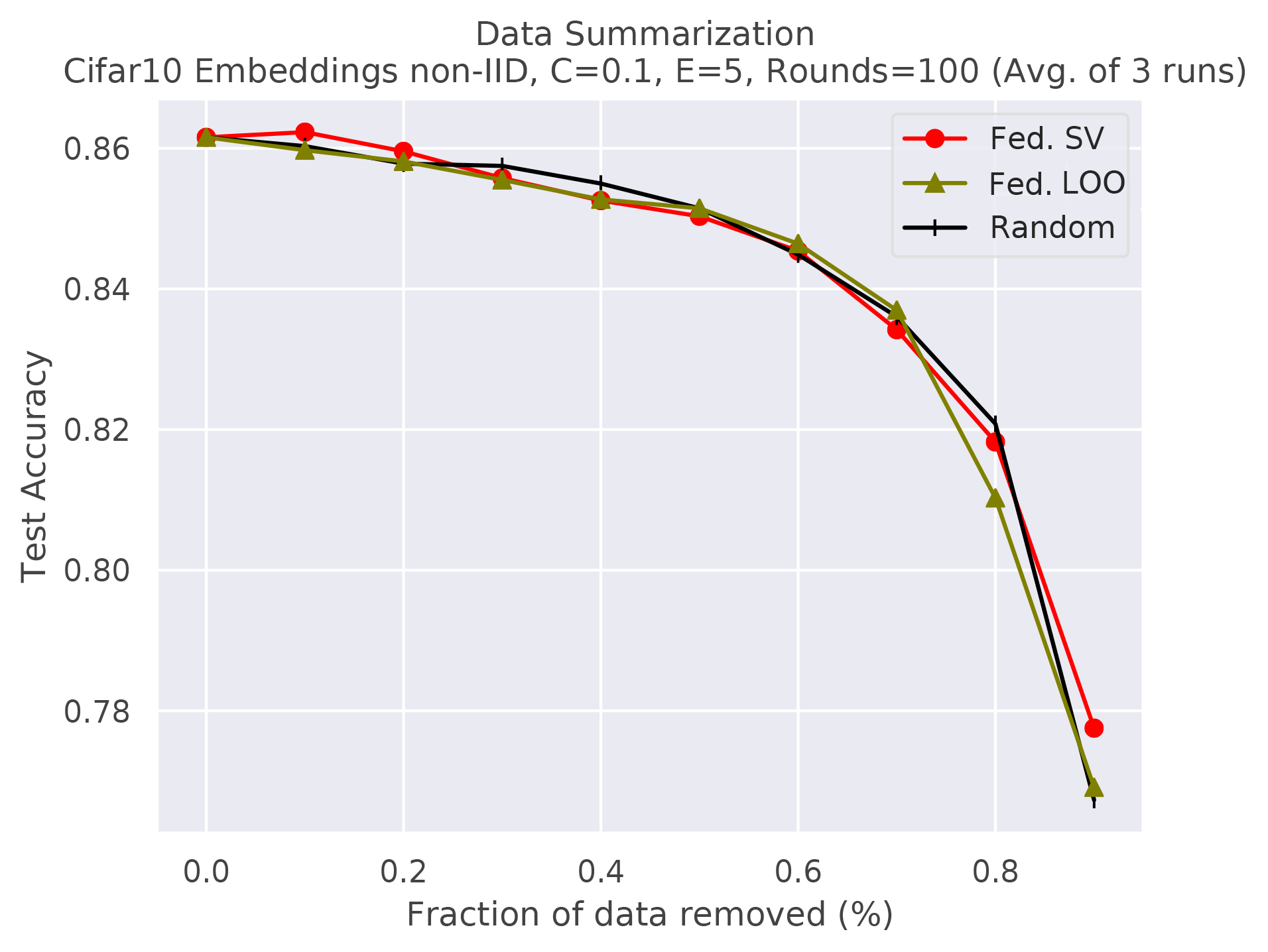}
  }
\caption{Data summarization experiments on  MNIST (top) and Cifar10 (bottom). }
\label{fig:mnist_summarization_10rounds_and_25rounds}
\end{figure}

\bibliography{main}
\bibliographystyle{abbrvnat}

\end{document}